\documentclass[conference]{IEEEtran}
\IEEEoverridecommandlockouts
\usepackage{cite}
\usepackage{amsmath,amssymb,amsfonts}
\usepackage{graphicx}
\usepackage{textcomp}
\usepackage{xcolor}
\usepackage{booktabs}
\usepackage{stfloats}
\usepackage{algorithm}
\usepackage{algpseudocode}
\usepackage{setspace}
\usepackage{hyperref}
\def\BibTeX{{\rm B\kern-.05em{\sc i\kern-.025em b}\kern-.08em
    T\kern-.1667em\lower.7ex\hbox{E}\kern-.125emX}}
\begin{document}

\title{FinXplore: An Adaptive Deep Reinforcement Learning Framework for Balancing and Discovering Investment Opportunities}
% FinXplore or XploreDRL or DRL with Exploration New Investment Opportunities (Exploration-Driven Portfolio Optimization using Deep Reinforcement Learning or A Novel Deep RL Framework for Discovering and Balancing Investment Opportunities

% \author{\IEEEauthorblockN{Anonymous Authors}}

\author{
	\IEEEauthorblockN{
		\begin{tabular}{c c}
			\begin{tabular}{c}
				1\textsuperscript{st} Himanshu Choudhary \\ 
				\textit{School of Mathematical $\&$ Statistical Sciences} \\
				\textit{Indian Institute of Technology Mandi, India} \\
				ch.himanshu1199@gmail.com
			\end{tabular} \vspace*{5mm}
			&
			\begin{tabular}{c}
				2\textsuperscript{nd} Arishi Orra \\
				\textit{School of Mathematical $\&$ Statistical Sciences} \\
				\textit{Indian Institute of Technology Mandi, India} \\
				arishi.orra98@gmail.com
			\end{tabular}
			\\
			\begin{tabular}{c}
				3\textsuperscript{rd} Manoj Thakur \\
				\textit{School of Mathematical $\&$ Statistical Sciences} \\
				\textit{Indian Institute of Technology Mandi, India} \\
				manoj@iitmandi.ac.in
			\end{tabular}
			&
		\end{tabular}
	}
}

%\author{\IEEEauthorblockN{1\textsuperscript{st} Himanshu Choudhary}
%\IEEEauthorblockA{\textit{School of Mathematical and Statistical Sciences} \\
%\textit{Indian Institute of Technology Mandi, India} \\
%ch.himanshu1199@gmail.com}
%\and
%\IEEEauthorblockN{2\textsuperscript{nd} Arishi Orra}
%\IEEEauthorblockA{\textit{School of Mathematical and Statistical Sciences} \\
%\textit{Indian Institute of Technology Mandi, India} \\
%arishi.orra98@gmail.com}
%\and
%\IEEEauthorblockN{3\textsuperscript{rd} Manoj Thakur}
%\IEEEauthorblockA{\textit{School of Mathematical and Statistical Sciences} \\
%\textit{Indian Institute of Technology Mandi, India} \\
%manoj@iitmandi.ac.in}
% \and
% \IEEEauthorblockA{\textit{School of Mathematical and Statistical Sciences} \\

\maketitle
%%%%%%%%%%%%%%%%%%%%%%%%%%%%%%%%%%%%%%%%%%%%%%%%%%%%%%%%%%%%%%%%%%%%%%%%%%%%%%%%%%%%%%%%%%%%%%%%%%%%
\begin{abstract}
    Portfolio optimization is essential for balancing risk and return in financial decision-making. Deep Reinforcement Learning (DRL) has stood out as a cutting-edge tool for portfolio optimization that learns dynamic asset allocation using trial-and-error interactions. However, most DRL-based methods are restricted to allocating assets within a pre-defined investment universe and overlook exploring new opportunities. This study introduces an investment landscape that integrates exploiting existing assets with exploring new investment opportunities in an extended universe. The proposed approach leverages two DRL agents and dynamically balances these objectives to adapt to evolving markets while enhancing portfolio performance. One agent allocates assets within the existing universe, while another assists in exploring new opportunities in the extended universe. The efficiency of the proposed methodology is determined using two real-world market data sets. The experiments demonstrate the superiority of the suggested approach against the state-of-the-art portfolio strategies and baseline methods.
\end{abstract}
%%%%%%%%%%%%%%%%%%%%%%%%%%%%%%%%%%%%%%%%%%%%%%%%%%%%%%%%%%%%%%%%%%%%%%%%%%%%%%%%%%%%%%%%%%%%%%%%%%%%
%%%%%%%%%%%%%%%%%%%%%%%%%%%%%%%%%%%%%%%%%%%%%%%%%%%%%%%%%%%%%%%%%%%%%%%%%%%%%%%%%%%%%%%%%%%%%%%%%%%%
\section{Introduction}

    Portfolio optimization is one of the essential problems in the modern financial domain. A portfolio is a combination of assets, investments, or funds. The diversification of the portfolio plays a vital role in mitigating the risk due to the dynamic market conditions. In 1952, Markowitz \cite{b1} laid the cornerstone of modern portfolio theory to tackle the portfolio optimization problem using mathematical and statistical tools. This broadly accepted theory provides a framework for building portfolios that successfully maintain a balance between return and risk. This concept later paved the way for subsequent research \cite{b2,b3,b4}. As a result of earlier breakthrough successes in gaming and robotics (e.g., AlphaGo, Atari see \cite{b5,b6}), reinforcement learning (RL) grabs the spotlight to solve dynamic decision-making problems like portfolio optimization \cite{b7,b8,b9, b10}. RL algorithms aim to learn the optimal allocations through end-to-end trial and error. A deep RL agent can be sufficiently flexible to identify the near-optimal allocations by interacting with the stock market environment. It combines exploration and exploitation and learns from the feedback received in terms of the rewards. A model free and policy based DRL model was effectively used by Jiang et al. \cite{b7} for managing cryptocurrency portfolios. In this study, the author established a framework that can be adapted to accommodate different variants of Deep Neural Networks. In addition, it can linearly scale the portfolio size by employing an Ensemble of Identical Independent Evaluators (EIIE) meta-topology. Afterward, Ye et al. \cite{b11} introduced a state-augmented reinforcement learning framework (SARL) for portfolio management that incorporates asset price predictions to handle data heterogeneity and market uncertainty. The framework is evaluated on Bitcoin and High-Tech stock markets, demonstrating superior performance over standard RL baselines.

    However, most existing works have fixed investment opportunities, i.e., one can only allocate all their wealth within a pre-defined investment universe. Investing in a fixed investment universe isolates investors from new or emerging investment options and dynamic market trends, which can adversely impact long-term returns and risk management. It would be better to create a balanced and diversified portfolio. Therefore, it is necessary to define a new investment landscape by combining the current universe with a novel exploration of new investment opportunities. By providing the option to search for new investment opportunities, the agent invests a portion of their wealth for exploration. The agent must deal with the trade-off between exploitation and exploration during this procedure. Wang et al. \cite{b12} proposed an approach for mean-variance (MV) portfolio optimization using RL. This approach attained the optimal balance between exploration and exploitation and an entropy-regularized reward function incorporated into the objective, improving exploitation. However, the authors do not consider the new investment opportunities, and it is unclear how exploration helps the agent. Afterwards, Aquino et al. \cite{b13} provided an extended investment universe model for portfolio optimization where an agent uses the MV model for exploitation and explores a new risky asset in the extended universe. This study describes a trade-off between exploration and exploitation for portfolio optimization, where the agent gains a concrete advantage from exploration by finding a new investment opportunity. However, the authors have used the classical mean-variance model for portfolio selection.
    
    This study uses deep RL agents for portfolio selection and explores new investment opportunities in an extended investment universe. In our approach, exploitation refers to investing in the already-existing universe of assets/funds, whereas exploration corresponds to searching for new investment opportunities to enhance the learning process and improve the policy. Our study aims to establish an infrastructure for portfolio managers or individual investors so that they can successfully figure out a balance between the exploitation of already-existing opportunities for investment and exploring new potential ones. We utilize two DRL agents, one to optimally allocate the assets within the existing universe and the other to explore new investment opportunities from the extended universe. To the best of our knowledge, no prior research has considered the potential for expanding the investing universe and using exploration to find new assets using deep RL agents. The main contribution of our study is as follows
    \begin{itemize}
        \item This study introduces an investment landscape for portfolio optimization that extends the investment universe to grab new or emerging investment opportunities.
        \item The proposed dual-agent architecture employs two DRL agents. Agent 1 optimizes portfolio weights within the existing investment universe while Agent 2 explores and suggests new assets to Agent 1 from the extended investment universe.
        \item The effectiveness of our proposed approach is demonstrated through empirical studies conducted on two major global stock markets. The findings highlight its ability to generate superior returns across various risk and return measures against the benchmarks.
    \end{itemize}

    The rest of the paper is organized as follows: Section \ref{Background} provides the background and the problem setup for the portfolio optimization task. Section \ref{Proposed} introduced the proposed FinXplore methodology in detail. The data description, experimental setup, and performance comparison of the proposed approach are reported in section \ref{Exp}. Finally, Section \ref{Conclusion} offers concluding remarks and suggests some future implications of our work.  
%%%%%%%%%%%%%%%%%%%%%%%%%%%%%%%%%%%%%%%%%%%%%%%%%%%%%%%%%%%%%%%%%%%%%%%%%%%%%%%%%%%%%%%%%%%%%%%%%%%%
\section{Background and Problem Setup} \label{Background}
    Portfolio optimization refers to the continuous redistributing of capital among various financial instruments and building a diversified portfolio. The objective is to achieve higher long-term cumulative returns while maintaining a tolerable level of risk. Due to its sequence-to-sequence learning capabilities, RL is a suitable candidate for continuous decision-making tasks, such as portfolio optimization. Markov Decision Process (MDP) provides an ideal mathematical framework for addressing RL problems. The portfolio optimization problem can be modeled as MDP and described as a tuple $(s, a, p, r, \gamma)$, where $s$ and $a$ represent the state and the action space, respectively, $r$ is the reward function defined as $r:s \times a \to \mathbb{R}$, $p$ is the state transition probability, and $\gamma$ is the discount factor varies between $0$ and $1$. The solution of the MDP is policy $\pi(a|s)$, which determines the action taken by the agent in any given state $s$. The agent aims to find the optimal policy $\pi^{*}$ that maximizes the expected discounted cumulative rewards, i.e., $\mathbb{E} \left[ \displaystyle \sum_{t=0}^{\infty} \gamma^{t} r(s_t, a_t) \right]$ in infinite time horizon settings by interacting with the environment. 
    
    \subsection{Environment for Portfolio Optimization} 
        The environment in an RL scenario is a representation of the surroundings in which the agent interacts and gains knowledge (learns). It is a composition of the state space, the action space, and the reward function. Our environment is motivated by an established architecture presented by \cite{b14}.   

        \textbf{State Space $s_t$:} A state of the environment consists of the relevant information at the time step $t$ that encompasses the market conditions, technical indicators, and other relevant factors. Specifically, the state space $s_t$ includes OHLCV (open-high-low-close-volume), daily price returns, the covariance matrix of the closing price of the assets, and eight technical indicators. ($30$ and $60$ day Simple Moving Averages (SMA), Moving Average Convergence Divergence (MACD), Upper and lower Bollinger bands, Relative Strength Index (RSI), Commodity Channel Index (CCI), and Average Directional Index (ADX)) corresponding to each asset.
    
        \textbf{Action Space $a_t$:} The actions are represented as a $(n+1)$ dimensional vector $w_t =[w_{1,t}, w_{2,t}, \ldots w_{n,t}, w_{\kappa,t} ] \in \mathbb{R}^{1 \times (n+1)}$. Where each component $w_{i,t}$ signifies the allocation of weight to asset $i$ in the existing investment universe, and $w_{\kappa,t}$ signifies the allocation of weight to the explored asset in the extended investment universe at the period $t \in \{1,2, \ldots, T\}$. In addition, we are ensuring that $\displaystyle \sum_{i=1}^{n} w_i + w_{\kappa} = 1$, with $0 \leq w_i, w_{\kappa} \leq 1$. To enforce these constraints, one can utilize the softmax activation function in the agent’s continuous actions.
    
        \textbf{Reward $r_t$:} The reward function incentivizes the agent to learn and refine its policy. This study uses the Sharpe Ratio (SR) \cite{b15}, a popular metric to access the risk-adjusted portfolio return, as a reward and can be defined as
        $$
            r_{t} = SR_{60}
        $$
        where $SR_{60}$ is the Sharpe Ratio calculated using the sliding window of the last $60$ portfolio returns. As defined by \cite{b16}, a transaction cost of $\delta = 0.05\%$ is assumed for executing the action $a_t$. This reward guides the agent to take the actions that maximize risk-adjusted returns rather than only returns.

        % $SR_{60} = \dfrac{\mu_{60}}{\sigma_{60}}$
 %        The portfolio return is the ratio of the portfolio value at time period $t$, $P_t$ to the portfolio value at time period $t-1$, $P_{t-1}$ and defined as
	% $$
	% R_{t} = w_{t-1} \cdot V_{t} - \delta \sum_{i=1}^{n} |w_{i,t} - w_{i,t-1}|
	% $$
	% where $V_{t} = \left[ V_{1,t}, V_{2,t}, \ldots , V_{n,t} \right]^T \in \mathbb{R}^{n \times 1}$, $V_{i,t} = \dfrac{P_{i,t}}{P_{i,t-1}} - 1$ for $i \in \left\{ 1,2, \ldots, n \right\}$ and $P_{i,t}$ is the price of asset $i$ at period $t$. $\delta$ is the transaction cost of executing action $a_t$. This reward measures the relative change in the portfolio value.
	
    \subsection{Deep Reinforcement Learning Agents}
        Deep Reinforcement Learning (DRL) agents fall into two major groups of algorithms: Model-based and Model-free. Model-based agents try to maximize the reward by performing some action regardless of the significance of the actions and utilizing a model of the environment's dynamics. Meanwhile, model-free agents, like Policy Gradient and Q-learning, aim to learn the optimal policy directly from experience.
        
        \subsubsection{Agent 1 - Proximal Policy Optimization (PPO)}
            Proximal Policy Optimization (PPO) is a new class of Policy Gradient (PG) algorithms proposed by \cite{b17}. A minor change in the network parameters of PG algorithms may lead to a significant change in the policy. PPO is a stochastic policy gradient algorithm that uses the clipped objective function and limits policy network updates to address the high sensitivity of PG algorithms to perturbations. PPO comprises two networks: the actor-network (AN) and the critic-network (CN). AN determines the optimal policy, while CN allows AN to get feedback and improve its decision-making process by estimating the value function and assessing the effectiveness of the policy. By clipping the probability ratio $J_t(\theta)$, which shows the difference between the current policy and the prior one, PPO improves the agent's training stability and guards against unnecessarily large policy updates. The objective function of PPO is defined as:
    	$$
    	   L( \theta ) = \hat{\mathbb{E}}_{t} \left[ \text{min}\left\{J_t(\theta) \cdot \hat{A_t}, \text{clip} \left(J_t(\theta), 1 - \epsilon, 1+\epsilon\right) \cdot \hat{A_t} \right\} \right]
    	$$
	    where $J_t(\theta) = \dfrac{\pi_{\theta}(a_t|s_t)}{\pi_{\theta_{old}}(a_t|s_t)}$. $J_t(\theta)$ is clipped within the range $\left[1-\epsilon,1+\epsilon\right]$, thereby discouraging significant deviations of the current policy from the prior one, and $\hat{A_t}$ is the advantage function, defined as $\hat{A}_{\pi}(s_t,a_t) = Q_{\pi}(s_t, a_t) - V_{\pi}(s_t)$, compares the effectiveness of action $a_t$ to the average of other actions taken at that particular state $s_t$.

        \subsubsection{Agent 2 - Deep Q-Learning (DQL)} 
            Q-learning is an effective tool for RL when dealing with discrete action spaces. However, its fundamental form could encounter difficulties in high-dimensional or continuous state spaces, which are frequently seen in financial applications. Mnih et al. \cite{b18} explored the concept of Deep Q-learning (DQL), an extension of Q-learning that substitutes a neural network for the Q-table to tackle these difficulties. The neural network $\theta$ is utilized to learn the weight for Q-values approximation, i.e., $Q(s,a,\theta)$, and choose the action $a$ with the highest Q-values in the given state $s$. The loss function $L(\theta)$ for training of DQN agent with parameter $\theta$ is defined as
            $$
            L(\theta) = \mathbb{E} \left[ \left( \hat{y} - Q(s_t, a_t; \theta) \right)^2 \right]
            $$
            where the neural network with parameter $\theta_{\text{target}}$ referred as the target Q-network which estimates the target values $\hat{y} = r_t + \gamma \max_{a} \hat{Q}(s_{t+1}, a_{t+1}; \theta_{\text{target}})$. Both the networks, the main network and the target network, are almost similar. The sole difference is that the target network is not updated at each step and freezes for some steps to prevent rapid oscillations in learning.

        % \subsubsection{TD3} TD3 algorithm enhances the DDPG algorithm by addressing its primary challenges. First, TD3 introduces target policy smoothing, where Gaussian noise is added to the target action:
        % $$a' = \mu_{\phi'}(s') + \epsilon, \quad \epsilon \sim \text{clip}(\mathcal{N}, -c, c)$$
        % This regularization technique helps prevent overfitting to sharp peaks in the Q-value estimates by smoothing the target policy. Second, this algorithm adopts clipped double Q-learning, utilizing two critic networks to reduce overestimation bias. The target value is calculated using the minimum output of the two critics:
        % $$Y = r + \gamma \min_{j=1,2} Q_{\theta'_j}(s', a')$$
        % Finally, TD3 employs delayed policy updates, where the actor network $\mu_\phi$, as well as the target actor and critic networks $\mu_{\phi'}$ and $Q_{\theta'_j}$, are updated less frequently than the critic networks $Q_{\theta_j}$. This strategy ensures that policy updates are made only after the Q-value estimates stabilize, resulting in more reliable learning.
%%%%%%%%%%%%%%%%%%%%%%%%%%%%%%%%%%%%%%%%%%%%%%%%%%%%%%%%%%%%%%%%%%%%%%%%%%%%%%%%%%%%%%%%%%%%%%%%%%%%
\section{Proposed Methodology} \label{Proposed}
    In standard portfolio optimization models, an agent allocates available wealth across a fixed investment universe comprising $n$ risky assets. The agent rebalances the portfolio at each period $t$ according to changing market conditions and reallocates the assets. However, investing in a fixed investment universe keeps investors away from dynamic market trends and new investment possibilities. Therefore, it is necessary to formalize an investment landscape by allowing the agent to devote some wealth $\kappa \geq 0$ to explore new investment opportunities in an extended universe $\mathcal{E}$.

\begin{figure*}[!htbp]
        \centering
        \includegraphics[width=\textwidth]{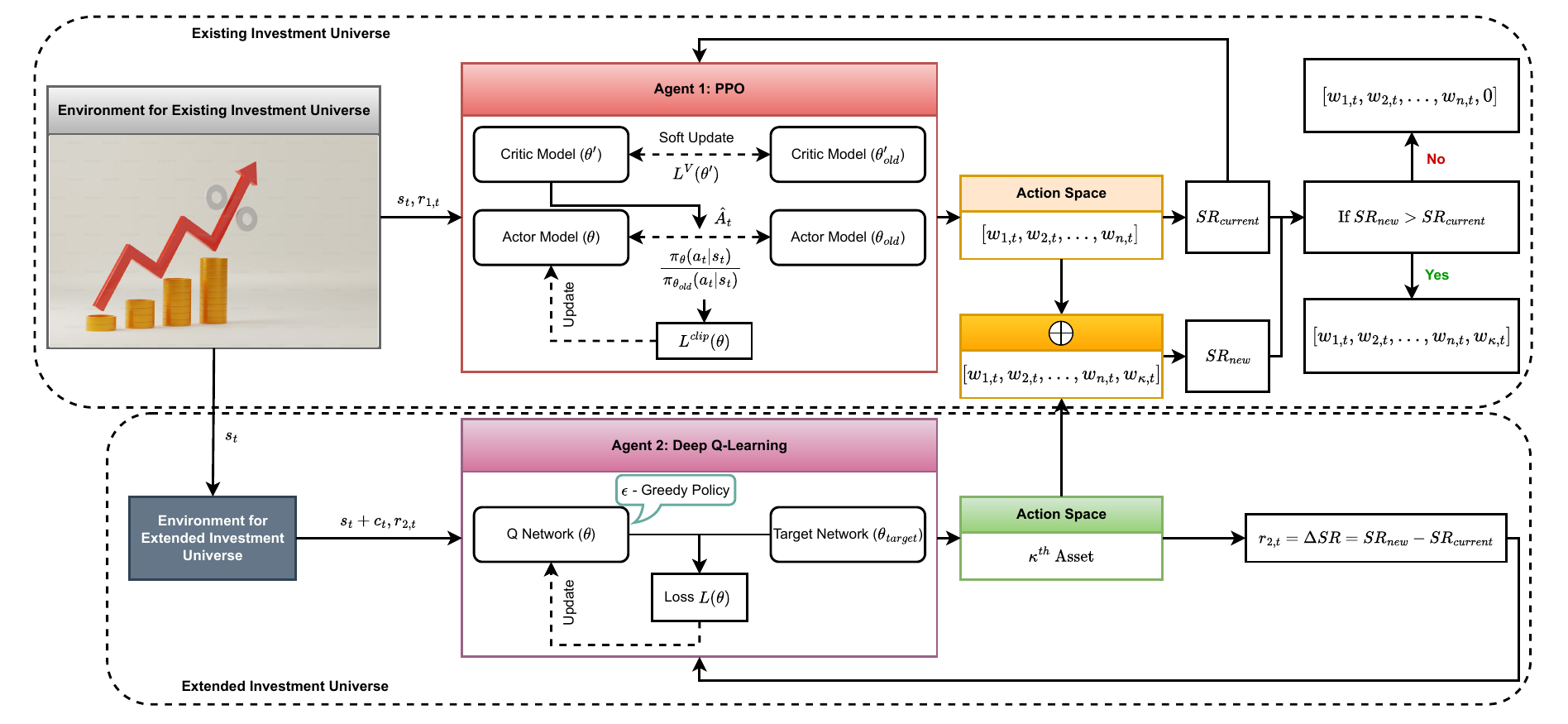} % Replace with your plot file
        \caption{The overall architecture of FinXplore}
        \label{fig:Proposed_Structure}
    \end{figure*}

    In this proposed study, we are utilizing two DRL agents: Agent 1 and Agent 2. Initially, Agent 1 observes the state from the environment, allocates the wealth across the existing investment universe, and provides the actions $w_i$ (portfolio weights). Agent 1 gets a reward based on the portfolio performance after weight rebalancing at the end of each period. Subsequently, Agent 1 reserves some fixed amount of $\kappa$ to explore new investment opportunities in the extended universe. For the exploration, we employ Agent 2 to identify new assets (stocks, commodities, Index Funds, etc.) from an extended investment universe $\mathcal{E}$. At each period, it recommends assets whose inclusion enhances the portfolio's performance. The explored assets exhibit low or negligible correlation with the existing asset universe, enabling the portfolio to diversify effectively and allocate capital to better-performing assets during periods of market underperformance. The complete proposed architecture is provided in Fig. \ref{fig:Proposed_Structure}. 

    Agent 2 observes the state, which appends the returns of the assets in $\mathcal{E}$ so that the state has both the current market situation with extended universe data and performs action $a^{*}$. Where $a^{*}$ represents the inclusion of an asset from the extended universe $\mathcal{E}$. Agent 2 gets a reward based on the marginal improvement of the portfolio's Sharpe Ratio:
    $$
        R_{\text{DQN}} = \Delta SR = SR_{\text{new}} - SR_{\text{current}}
    $$
    Where $SR_{\text{new}}$ is the Sharpe Ratio after adding the asset suggested by Agent 2 and $SR_{\text{current}}$ is the Sharpe Ratio of the existing portfolio. Thus, Agent 2 is incentivized proportional to how much it improves the portfolio's performance. Agent 1 accepts or rejects Agent 2's suggestion based on whether the suggested asset improves the portfolio performance and re-optimizes the asset weights accordingly. This feedback helps Agent 2 learn which assets (or asset features) are more likely to contribute positively to the portfolio. If Agent 2 successfully selects an asset that increases the portfolio's Sharpe Ratio, Agent 1 allocates $\kappa\%$ of total wealth into this asset and invests the remaining wealth into the existing universe. On the contrary, Agent 1 allocates all the wealth to the existing universe. Agent 2 learns to select assets more likely to improve SR, and Agent 1 learns to optimize weights effectively, given these assets. Both agents achieve higher rewards by working independently. Over time, both agents learn policies that complement each other, improving portfolio performance. The complete training algorithm is provided in Algorithm \ref{alg:combined}.
    \begin{algorithm}
    \caption{Portfolio Optimization with Exploration Assistance}
    \label{alg:combined}
    \begin{algorithmic}[1]
        \Require States $s_t$ and $c_t$ for Existing and Extended Investment Universes respectively, Agent 1 (PPO), Agent 2 (DQN), Existing Universe $\mathcal{U}$, Extended Universe $\mathcal{E}$
        \Ensure Updated portfolio weights $w$
        
        \State $a_{\mathcal{U}} \leftarrow \text{PPO}(s_t, a_t)$ Portfolio weights from Agent 1
        \State $SR_{current} \leftarrow f(a_{\mathcal{U}})$ Sharpe Ratio for current portfolio
        
        \State $a_{*} \leftarrow \text{DQN}(s_t + c_t, \mathcal{E})$ Agent 2 selects asset from extended universe $\mathcal{E}$
        \State $a_{\mathcal{E}} \leftarrow \text{Reoptimize Portfolio} (a_{\mathcal{U}}, a^*)$ 
        \State $SR_{new} \leftarrow f(a_{\mathcal{E}})$ Sharpe Ratio after adding new asset
        
        \If{$SR_{new} > SR_{current}$}
            \State $w \leftarrow a_{\mathcal{E}}$
        \Else
            \State $w \leftarrow a_{\mathcal{U}}$
        \EndIf
        \State $R_{\text{PPO}} \leftarrow SR_{current}$
        \State $R_{\text{DQN}} \leftarrow \Delta SR = SR_{new} - SR_{current}$
        \State Update PPO and DQN policies using $R_{\text{PPO}}$ and $R_{\text{DQN}}$.
    
    \end{algorithmic}
\end{algorithm}

% This means that the agent updates the portfolio weights at each period to reflect the asset allocation after interacting with the environment.
%%%%%%%%%%%%%%%%%%%%%%%%%%%%%%%%%%%%%%%%%%%%%%%%%%%%%%%%%%%%%%%%%%%%%%%%%%%%%%%%%%%%%%%%%%%%%%%%%
%%%%%%%%%%%%%%%%%%%%%%%%%%%%%%%%%%%%%%%%%%%%%%%%%%%%%%%%%%%%%%%%%%%%%%%%%%%%%%%%%%%%%%%%%%%%%%%%%%%%
\section{Experiment} \label{Exp}
    
    \subsection{Data Description}
        The efficiency of the suggested methodology is evaluated using the two well-known global market indices: the NIFTY Index of the NSE Mumbai, India and the DJIA Index of the NYSE New York, USA. For the existing investment universe, we selected the daily historical data of $18$ randomly chosen stocks for each market instance from January $2011$ to November $2024$. The data set comprises open, high, low, and closed (OHLC) prices collected from Yahoo Finance. For the extended investment universe, we selected the daily historical data of the five financial instruments: Gold, crude oil, silver, copper and natural gas for the same period. The first $11$ years of data from January $2011$ to December $2021$ is used to train the agents, and we back-test the proposed methodology on the latest data from January $2022$ to November $2024$. For both the market indices, we adopt similar training and trading strategies.
        
    \subsection{Experimental Setup}
        Carefully adjusting the hyperparameters of the DRL agent is often required to improve their performance. We utilized the \href{https://hyperopt.github.io/hyperopt/}{Hyperopt} Python module for hyperparameter optimization. Table \ref{Exp_Para} provides the range of hyperparameters chosen for the DRL agents based on empirical studies \cite{b16, b17, b18, b19}. 
        % The agent begins trading with an initial capital of $1$ million.
	
	\begin{table}[!htp]\centering
		\caption{Parameters setting for experiment}\label{Exp_Para}
		\scriptsize
		\begin{center}
			\begin{tabular}{cccccc} \toprule
				\textbf{Parameter}              &\textbf{Range}         \\ \midrule
				Hidden Layers               & $[1,8]$               \\
				Hidden Layers Dim           & $[2,512]$             \\
				Learning rate               & $[10^{-8}, 10^{-1}]$  \\
				Discount factor ($\gamma$)  & $[0,1]$               \\
				Activation Function         & [Relu, Sigmoid, Tanh] \\
				Dropout rate                & $[0,0.5]$             \\
				Entropy coefficient         & $[0.01,0.1]$          \\
				Value function coefficient  & $[0.5,1]$             \\
                    $\epsilon$-clip             & $0.2$                 \\
				PPO epochs                  & $[5,50]$              \\
                    Batch size for Q-network    & $[32, 256]$           \\
				No. of episodes             & $500$                 \\
                    $\kappa$                    & $10\%$                 \\
				\bottomrule
			\end{tabular}
		\end{center}
	\end{table}
    
    \subsection{Baseline Strategies}
        We compare our proposed methodology with the following benchmark strategies: \\
        \textbf{DRL Agents:} The proposed methodology is compared with the DRL agent without exploring the new investment opportunities in the extended universe. In addition, the proposed approach is compared with recent reinforcement learning methods for portfolio optimization, including EIIE \cite{b7} and SARL \cite{b11}. \\
        \textbf{Markowitz's Mean-Variance Optimization (MVO):} The MVO model attempts to maximize portfolio returns while minimizing portfolio risk, particularly portfolio volatility, and offers a mathematical framework for determining the optimal allocation. \\
        \textbf{Follow the Winner:} This strategy reallocates all portfolio weights to the stock with the highest return in the previous period. It mimics the behavior of typical investors, indicating that this stock will continue to perform well in the present period. \\
        \textbf{Follow the Loser:} This strategy behaves opposite to the follow-the-winner strategy and reallocates all portfolio weights to the stock with the lowest return in the previous period. It follows the concept that a strategy with poor past performance has a strong possibility of recovering. \\
        \textbf{Market Index:} A market index is a price-weighted index comprising the prominent stocks of the exchange, offering a reflection of the broader market. The Nifty $50$ and Dow $30$ indices served as benchmarks for the NIFTY and DJIA stock markets.
        
    \subsection{Performance Metrics}
        The effectiveness of the proposed methodology is assessed using the following six performance metrics. \\
        \textbf{Cumulative Return:} The total returns achieved at the end of the trading period.  \\
        \textbf{Annual Return:} Annual return measures the percentage change in the returns accumulated over one year. \\
        \textbf{Sharpe Ratio:} It measures the risk-adjusted returns and is defined as the excess return generated by the trader per unit risk. \\
        \textbf{Calmar Ratio:} A risk-adjusted performance metric that measures the portfolio returns relative to the maximum drawdown. \\
        \textbf{Annual Volatility:} It calculates the risk of an investment and evaluates the yearly dispersion of returns. \\
        \textbf{Maximum Drawdown:} It is the highest recorded loss from any peak to a trough and assesses the portfolio's downside risk.
    
    \subsection{Results and Discussion}
        We backtest our proposed FinXplore approach and benchmark method for the trade period spanning January $1, 2022$, to November $30, 2024$. At the start of the trading period, an initial capital of $1$ million is allocated to the agent. The comparison of the performance of the proposed approach with benchmarks on the NIFTY and DJIA datasets is summarized in Table \ref{Tab1:Nifty} and Table \ref{Tab2:Dow}, respectively. The experiments were repeated five times, and the tables report the mean performance metrics along with their standard deviations across these runs. The highlighted text in bold indicates the best results. All the experiments are done in the same environment settings to ensure consistency in the results.

        A detailed performance comparison of the proposed approach with the benchmarks on the NIFTY dataset is reported in Table \ref{Tab1:Nifty}. The findings show that FinXplore achieved the highest cumulative and annualized returns of $127.91\%$ and $33.53\%$, respectively, among all benchmarks. The proposed approach produced approx $3.4$ times higher returns than the NIFTY $50$ index, which provided a total return of $37.61\%$ throughout the trading period. The proposed FinXplore approach also demonstrated its superiority when considering risk-adjusted returns. While maintaining the highest Sharpe and Calmar ratios of $1.83$ and $2.06$, our proposed methodology far outperforms all other benchmarks. The DRL agent without exploration is the second-best performer in terms of Sharpe and Calmar ratios, closely followed by the SARL, EIIE, and Markowitz model. The volatility and maximum drawdown measure the risk of a portfolio during backtesting. The suggested approach has the lowest maximum drawdown closely followed by the market index. However, the Markowitz model is the least risky compared to all other models and maintains a lower volatility of $14.26\%$. The follow-the-loser and follow-the-winner strategies have the highest annual volatility and maximum drawdown, making them very risky. FinXplore exhibits superior performance in five out of six metrics, except for the volatility. The above findings highlight the robustness and effectiveness of our proposed approach.
        
    \begin{table*}[!htp]
		\centering
		\caption{Performance indicators of the Proposed FinXplore approach and Benchmarks for entire trading period on Nifty dataset} \label{Tab1:Nifty}
		\resizebox{\textwidth}{!}{%
			\begin{tabular}{lccccccccccc}\toprule
				% \textbf{Model/Benchmark} &\textbf{Comulative Return} &\textbf{Sharpe Ratio} &\textbf{Annual Volatility} &\textbf{Calmar Ratio} &\textbf{Sortino Ratio} &\textbf{Stability}  \\\midrule
				\textbf{Model/Benchmark} & \textbf{Cumulative Return (\%)} & \textbf{Annualized Return (\%)} & \textbf{Sharpe Ratio} & \textbf{Calmar Ratio} & \textbf{Annual Volatility (\%)} & \textbf{Maximum Drawdown (\%)} \\ \midrule
				\textbf{FinXplore}              & \textbf{127.91 $\pm$ 8.12}  & \textbf{33.53 $\pm$ 3.10}   & \textbf{1.83 $\pm$ 0.14}   & \textbf{2.06 $\pm$ 0.10} & 16.55 $\pm$ 1.25 & \textbf{16.31 $\pm$ 1.49}  \\
				Without Exploration      & 93.86 $\pm$ 8.35            & 26.15 $\pm$ 3.31            & 1.56 $\pm$ 0.14   & 1.30 $\pm$ 0.10          & 15.73 $\pm$ 1.42          & 20.19 $\pm$ 1.52          \\
				SARL      & 89.50 $\pm$ 7.23            & 24.65 $\pm$ 2.98            & 1.51 $\pm$ 0.13   & 1.24 $\pm$ 0.11          & 16.52 $\pm$ 2.02          & 20.68 $\pm$ 1.29          \\
				EIIE      & 82.07 $\pm$ 9.84            & 22.94 $\pm$ 4.05            & 1.46 $\pm$ 0.15   & 1.33 $\pm$ 0.12          & 14.90 $\pm$ 2.55          & 19.24 $\pm$ 1.35          \\
                    Markowitz                & 75.82                         & 21.90                         & 1.46                & 1.19                       & \textbf{14.26}                     & 18.37                       \\
                    Follow the Winner        & 74.96                         & 21.69                         & 0.80                & 0.49                       & 30.25                   & 44.39    \\
                    Follow the Loser         & 4.48                          & 1.55                          & 0.22                & 0.03                       & 36.64                   & 56.51    \\
				Nifty Index              & 37.61                         & 11.86                         & 0.83               & 0.71                       & 14.94                   & 16.71     \\
				\bottomrule
			\end{tabular}
		}
	\end{table*}

    Table \ref{Tab2:Dow} presents the performance measures of the proposed methodology on the DJIA dataset. Our agent outperformed benchmark models with a substantial margin and delivered cumulative and annual returns twice as high as the market index. Like the Nifty $50$ market, with the superior Sharpe and Calmar ratios of $0.90$ and $0.67$, the proposed approach efficiently generates risk-adjusted returns. However, the DRL agents showed high levels of annualized volatility and drawdowns, indicating significant risk. The DJIA market index and Markowitz model achieve the minimum volatility and maximum drawdown, respectively, slightly better than the proposed approach. All these performance metrics affirm that our proposed FinXplore approach produces outstanding returns at a tolerable risk.

    \begin{table*}[!htp]
		\centering
		\caption{Performance indicators of the Proposed FinXplore approach and Benchmarks for entire trading period on DJIA dataset} \label{Tab2:Dow}
		\resizebox{\textwidth}{!}{%
			\begin{tabular}{lccccccccccc}\toprule
				% \textbf{Model/Benchmark} &\textbf{Comulative Return} &\textbf{Sharpe Ratio} &\textbf{Annual Volatility} &\textbf{Calmar Ratio} &\textbf{Sortino Ratio} &\textbf{Stability}  \\\midrule
				\textbf{Model/Benchmark} & \textbf{Cumulative Return (\%)} & \textbf{Annualized Return (\%)} & \textbf{Sharpe Ratio} & \textbf{Calmar Ratio} & \textbf{Annual Volatility (\%)}  & \textbf{Maximum Drawdown (\%)}  \\ \midrule
				\textbf{FinXplore}              & \textbf{49.56 $\pm$ 3.82}   & \textbf{14.86 $\pm$ 1.85}   & \textbf{0.90 $\pm$ 0.08}   & \textbf{0.67 $\pm$ 0.06} & 16.98 $\pm$ 1.34 & 22.30 $\pm$ 1.46   \\
				Without Exploration      & 37.15 $\pm$ 4.06            & 11.49 $\pm$ 2.01            & 0.75 $\pm$ 0.09   & 0.52 $\pm$ 0.05          & 16.35 $\pm$ 1.38          & 21.90 $\pm$ 1.62    \\
				SARL      & 27.69 $\pm$ 3.21            & 8.79 $\pm$ 1.59            & 0.64 $\pm$ 0.08   & 0.43 $\pm$ 0.05          & 15.97 $\pm$ 1.22          & 20.60 $\pm$ 1.40          \\
				EIIE      & 25.78 $\pm$ 4.51            & 8.23 $\pm$ 2.22            & 0.61 $\pm$ 0.10   & 0.42 $\pm$ 0.07          & 14.44 $\pm$ 1.45          & 19.25 $\pm$ 1.70          \\
                    Markowitz                & 16.79                         & 5.49                          & 0.45                & 0.30                       & 13.94                     & \textbf{18.42}    \\
                    Follow the Winner        & 5.89                          & 1.99                          & 0.22                & 0.06                       & 32.11                     & 30.83    \\
                    Follow the Loser         & -23.27                        & -8.72                         & -0.13               & -0.13                      & 31.40                     & 67.89    \\
				Dow Index              & 23.23                         & 7.46                          & 0.59                & 0.34                       & \textbf{13.88}                     & 21.69     \\
				\bottomrule
			\end{tabular}
		}
	\end{table*}

\begin{figure*}[!htbp]
        \centering
        \includegraphics[width=\textwidth]{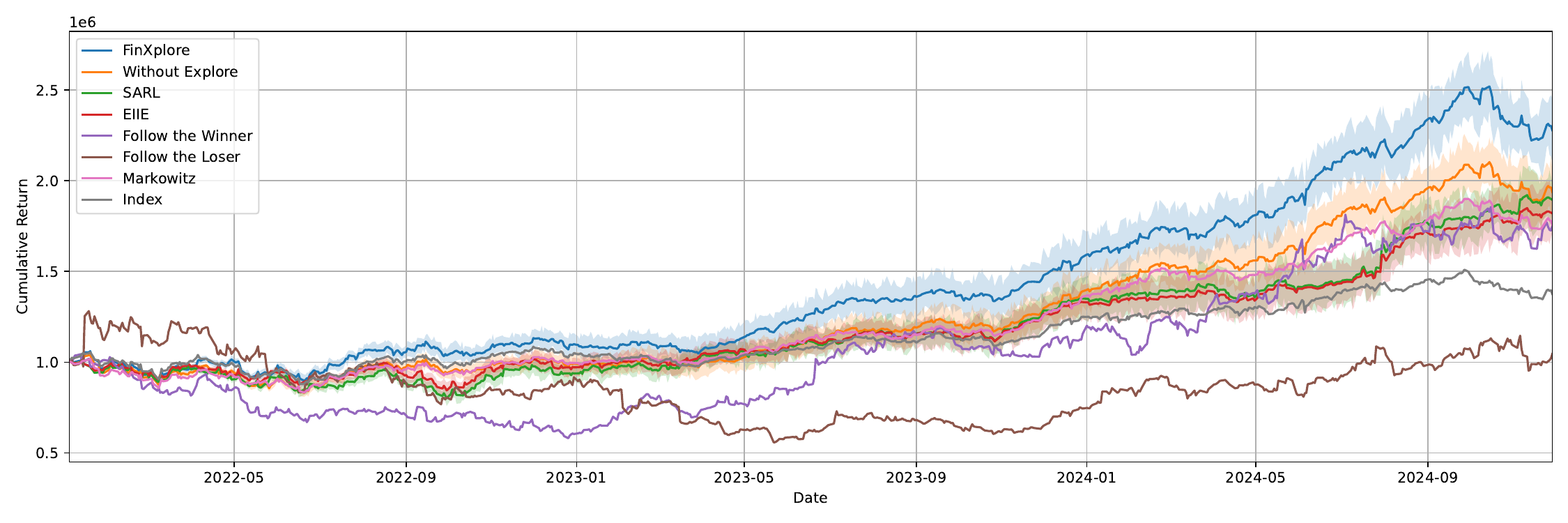} % Replace with your plot file
        \caption{Cumulative Return Plot of the proposed FinXplore approach and state-of-the-art portfolio strategies over the trading  period on Nifty Dataset}
        \label{fig:Nifty_Cum}
    \end{figure*}

\begin{figure*}[!htbp]
        \centering
        \includegraphics[width=\textwidth]{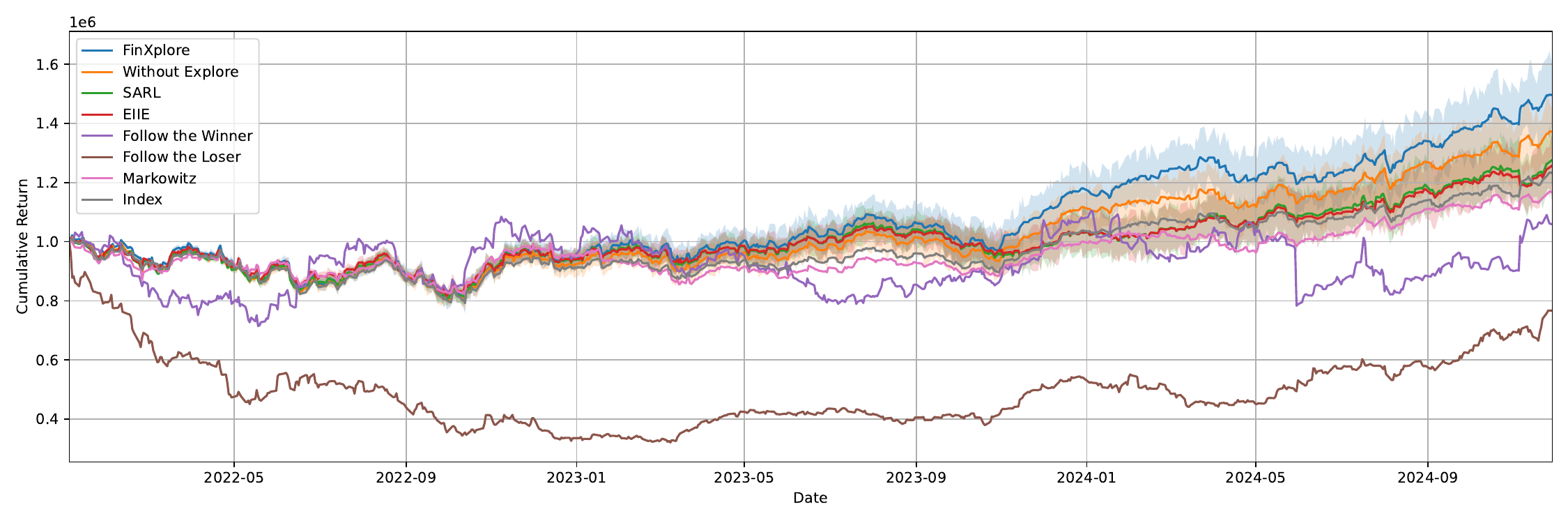} % Replace with your plot file
        \caption{Cumulative Return Plot of the proposed FinXplore approach and state-of-the-art portfolio strategies over the trading period on DJIA Dataset}
        \label{fig:Dow_Cum}
    \end{figure*}

    Fig. \ref{fig:Nifty_Cum} and Fig. \ref{fig:Dow_Cum} present the cumulative wealth plot of all models for the trading period on the NIFTY and DJIA datasets, respectively. Fig. \ref{fig:Nifty_Cum} illustrates that the market showed a bearish trend during the first two quarters, and all the models started losing wealth. However, the proposed approach showed little decline, demonstrating its risk-averse nature. Until October $2023$, FinXplore continuously made modest profits over the sideways market trend, making up for previous losses and building wealth. The market is bullish in the subsequent trading period, and all agents accumulate significant wealth. At the end of the trading period, the lower bound of the proposed approach outperforms the other benchmarks followed by DRL agent without exploration, SARL, EIIE, Markowitz, follow-the-winner, market index, and follow-the-loser, respectively. These results validated the efficacy of the proposed approach on the NIFTY dataset. In a similar way, Fig. \ref{fig:Dow_Cum} represents that our recommended approach exhibited dominating performance over the benchmarks on the DJIA dataset. However, our model struggles during the initial trading phase, and the follow-the-winner strategy has surpassed it for some time. But in the long run, the proposed method leaves behind this strategy with significant margins. The plot indicates that the FinXplore outshines their stability, outperforming Markowitz, the market index, and other benchmarks.

    % Comparison of the proposed approach with baselines strategies based on the cumulative wealth for the trading period without transaction cost.
    % Comparison of the proposed RA-DRL with base DRL agents and benchmarks based on the cumulative wealth for the trading period on Sensex
    % Cumulative returns of the proposed DREB model and the base DRL agents for the DJI data over the entire trading period.

\begin{figure*}[!htbp]
        \centering
        \includegraphics[width=\textwidth]{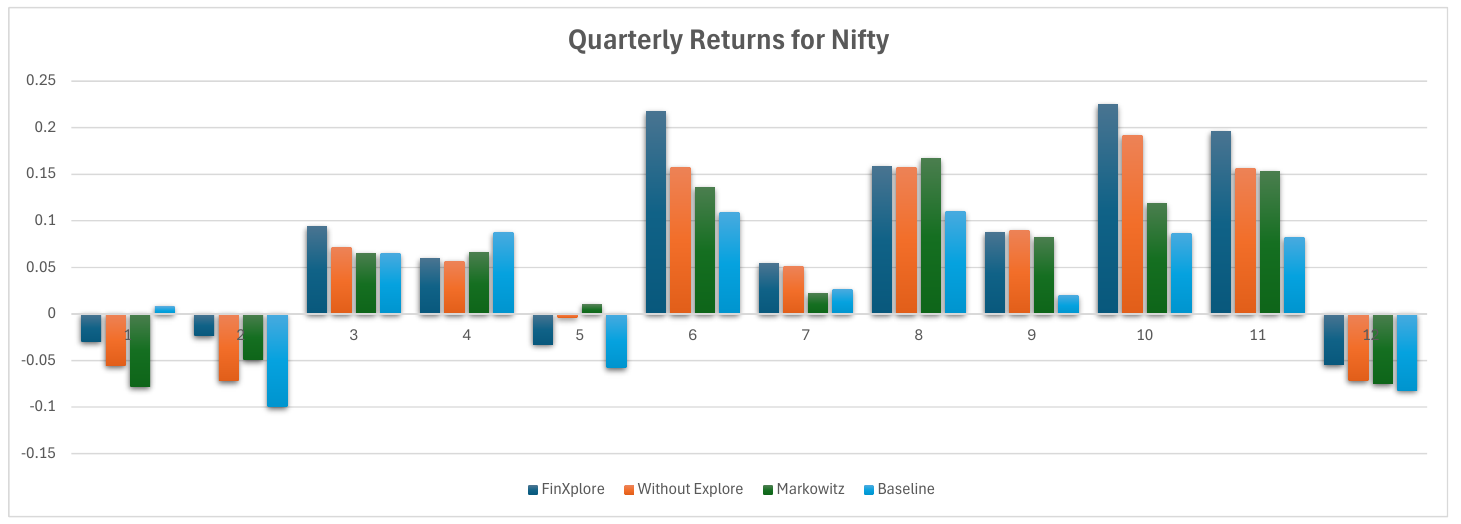} % Replace with your plot file
        \caption{Quarterly Returns for Nifty dataset over the trading period}
        \label{fig:Nifty_Quarter}
    \end{figure*}

\begin{figure*}[!htbp]
        \centering
        \includegraphics[width=\textwidth]{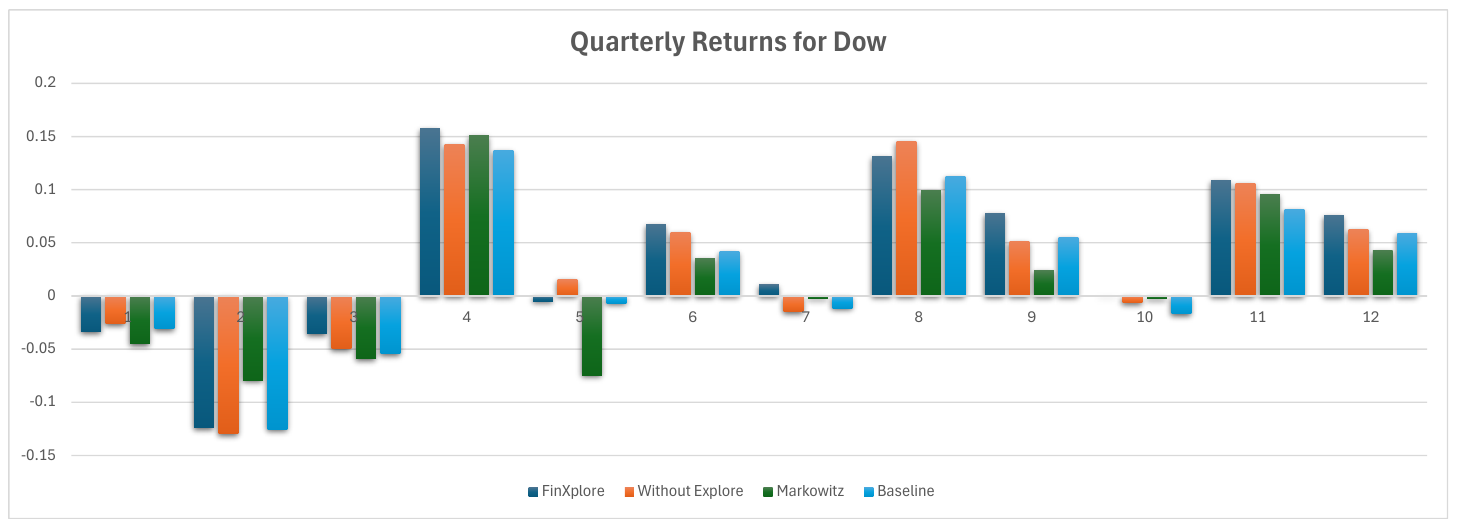} % Replace with your plot file
        \caption{Quarterly Returns for DJIA dataset over the trading period}
        \label{fig:Dow_Quarter}
    \end{figure*}

    Fig. \ref{fig:Nifty_Quarter} and Fig. \ref{fig:Dow_Quarter} showcase the quarterly returns generated by the proposed approach and benchmarks for the NIFTY and DJIA datasets, respectively. In most quarters, the suggested FinXplore approach regularly achieves noticeably better returns for both the datasets. However, our model displays comparatively lower or even negative returns in few quarters. This implies that although exploration has the potential to yield higher profits, it may also cause considerable volatility in certain market circumstances.

    % \begin{figure}[!htbp]
    %     \centering
    %     \includegraphics[scale=0.35]{quarterly_returns_Plot_Dow.pdf} % Replace with your plot file
    %     \caption{Quarterly Returns for Dow}
    %     \label{fig:single_column_plot}
    % \end{figure}
%%%%%%%%%%%%%%%%%%%%%%%%%%%%%%%%%%%%%%%%%%%%%%%%%%%%%%%%%%%%%%%%%%%%%%%%%%%%%%%%%%%%%%%%%%%%%%%%%%%%
\section{Conclusion and Future Scope} \label{Conclusion}
    This study introduces a novel investment landscape  FinXplore for portfolio optimization, which effectively combines asset allocation in an existing investment universe with the exploration of new investment opportunities in an extended universe. The proposed dual-agent architecture leverages the complementary strengths of two DRL agents to achieve superior portfolio performance. Both agents help each other to learn optimal policies, and this collaboration allows for a dynamic and adaptive approach to portfolio management. FinXplore performs noticeably better than benchmark strategies across key performance metrics on NIFTY and DJIA data sets. The findings emphasize the importance of including exploration in portfolio optimization. FinXplore offers a cutting-edge solution for portfolio managers and investors as it helps agents adjust to changing market conditions and uncover new investment possibilities.
    
    Future research could refine the framework by incorporating regulatory constraints and additional asset classes to enhance real-world applicability. The sentiment analysis and financial reports could improve the agent’s decision-making. Subsequent work could incorporate risk measures like Var and CVar in rewards to make the audience more aware of risk.
    % This study provides a solid and flexible solution for volatile financial markets, representing a major advancement in portfolio optimization. 
%%%%%%%%%%%%%%%%%%%%%%%%%%%%%%%%%%%%%%%%%%%%%%%%%%%%%%%%%%%%%%%%%%%%%%%%%%%%%%%%%%%%%%%%%%%%%%%%%%%%
% \section*{Acknowledgment}

% The preferred spelling of the word ``acknowledgment'' in America is without 
% an ``e'' after the ``g''. Avoid the stilted expression ``one of us (R. B. 
% G.) thanks $\ldots$''. Instead, try ``R. B. G. thanks$\ldots$''. Put sponsor 
% acknowledgments in the unnumbered footnote on the first page.
%%%%%%%%%%%%%%%%%%%%%%%%%%%%%%%%%%%%%%%%%%%%%%%%%%%%%%%%%%%%%%%%%%%%%%%%%%%%%%%%%%%%%%%%%%%%%%%%%%%%

%%%%%%%%%%%%%%%%%%%%%%%%%%%%%%%%%%%%%%%%%%%%%%%%%%%%%%%%%%%%%%%%%%%%%%%%%%%%%%%%%%%%%%%%%%%%%

\end{document}